\documentclass[10pt,twocolumn,letterpaper]{article}
\pdfoutput=1
\usepackage{cvpr}
\usepackage{times}
\usepackage{epsfig}
\usepackage{graphicx}
\usepackage{amsmath}
\usepackage{amssymb}
\usepackage{xcolor}
\usepackage{multicol}
\usepackage{gensymb}
\usepackage{booktabs}
\usepackage{multirow}
\usepackage{subcaption}

\usepackage[breaklinks=true,bookmarks=false]{hyperref}

\cvprfinalcopy 

\def\cvprPaperID{7} 

\title{Sensor Fusion for Joint 3D Object Detection and Semantic Segmentation}

\author{Gregory P. Meyer, Jake Charland, Darshan Hegde, Ankit Laddha, Carlos Vallespi-Gonzalez\\
Uber Advanced Technologies Group\\
{\{\tt\small gmeyer,jakec,darshan.hegde,aladdha,cvallespi\}@uber.com}}

\ifcvprfinal\fi
\begin{document}

\maketitle
\thispagestyle{empty}

\begin{abstract}
In this paper, we present an extension to LaserNet, an efficient and state-of-the-art LiDAR based 3D object detector. We propose a method for fusing image data with the LiDAR data and show that this sensor fusion method improves the detection performance of the model especially at long ranges. The addition of image data is straightforward and does not require image labels. Furthermore, we expand the capabilities of the model to perform 3D semantic segmentation in addition to 3D object detection. On a large benchmark dataset, we demonstrate our approach achieves state-of-the-art performance on both object detection and semantic segmentation while maintaining a low runtime.
\end{abstract}

\section{Introduction}

3D object detection and semantic scene understanding are two fundamental capabilities for autonomous driving. LiDAR range sensors are commonly used for both tasks due to the sensor's ability to provide accurate range measurements while being robust to most lighting conditions. In addition to LiDAR, self-driving vehicles are often equipped with a number of cameras, which provide dense texture information missing from LiDAR data. Self-driving systems not only need to operate in real-time, but also have limited computational resources. Therefore, it is critical for the algorithms to run in an efficient manner while maintaining high accuracy.

Convolutional neural networks (CNNs) have produced state-of-the-art results on both 3D object detection \cite{ibm,lasernet} and 3D point cloud semantic segmentation \cite{squeezeseg, zhang2018efficient} from LiDAR data.
Typically, previous work \cite{avod,ibm,hdnet,pixor,zhang2018efficient,voxelnet} discretizes the LiDAR points into 3D voxels and performs convolutions in the bird's eye view (BEV).
Only a few methods \cite{velofcn,lasernet,squeezeseg} utilize the native range view (RV) of the LiDAR sensor.
In terms of 3D object detection, BEV methods have traditionally achieved higher performance than RV methods.
On the other hand, RV methods are usually more computationally efficient because the RV is a compact representation of the LiDAR data where the BEV is sparse.
Recently, \cite{lasernet} demonstrated that a RV method can be both efficient and obtain state-of-the-art performance when trained on a significantly large dataset.
Furthermore, they showed that a RV detector can produce more accurate detections on small objects, such as pedestrians and bikes.
Potentially, this is due to the BEV voxelization removing fine-grain details which is important for detecting smaller objects.

\begin{figure}[t]
\centering
\includegraphics[width=0.43\textwidth]{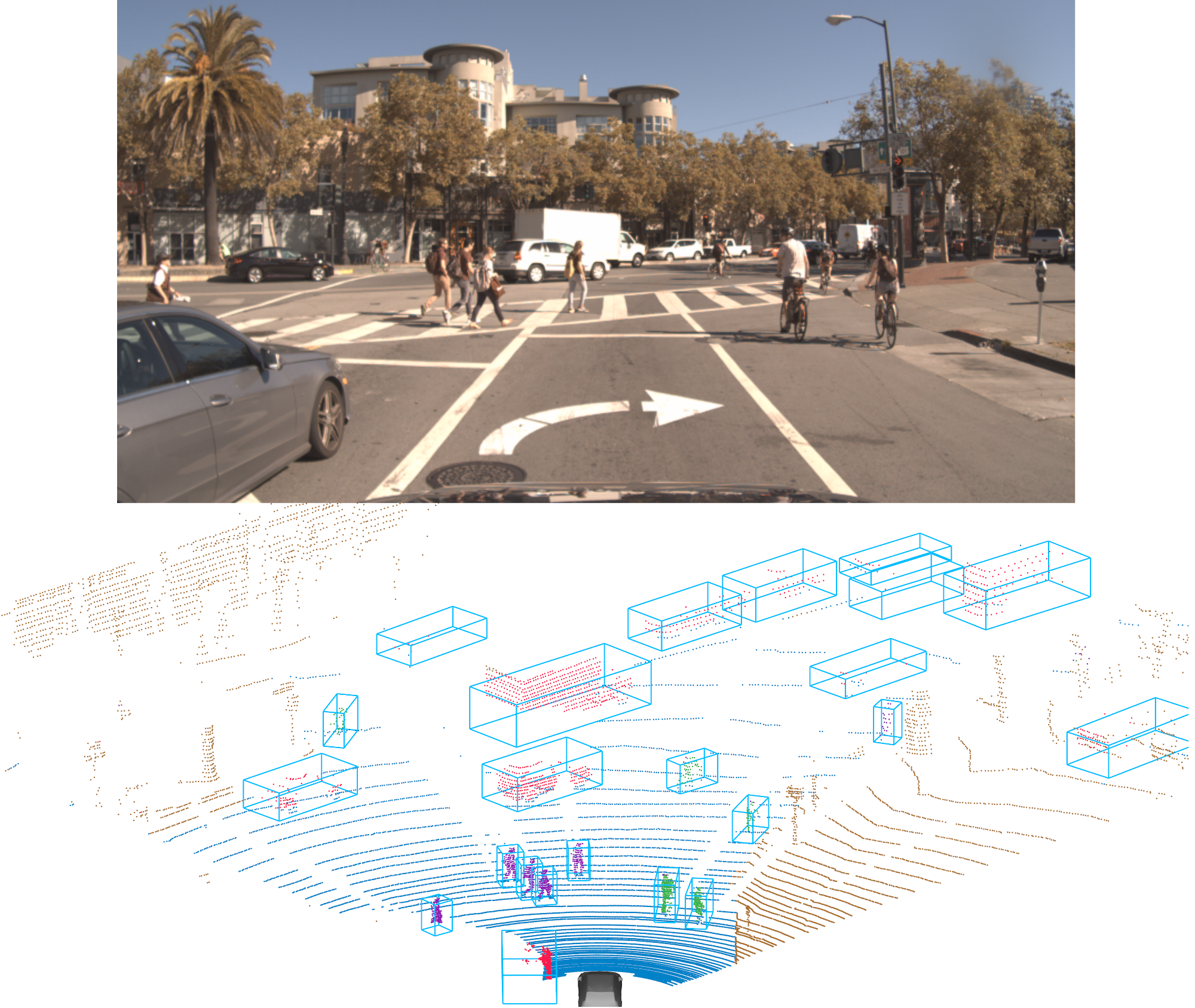}
\caption{Example object detection and semantic segmentation results from our proposed method. Our approach utilizes both 2D images (top) and 3D LiDAR points (bottom).}
\label{fig:example}
\end{figure}

At range, LiDAR measurements become increasingly sparse, so incorporating high resolution image data could improve performance on distant objects.
There have been several methods proposed to fuse camera images with LiDAR points \cite{mv3d,avod,ibm,cvpr17mousavian,frustum_pointnet,pointfusion}.
Although these methods achieve good performance, they are often computationally inefficient, which makes integration into a self-driving system challenging.

In this paper, we propose an efficient method for fusing 2D image data and 3D LiDAR data, and we leverage this approach to improve LaserNet, an existing state-of-the-art LiDAR based 3D object detector \cite{lasernet}.
Our sensor fusion technique is efficient allowing us to maintain LaserNet's low runtime.
Unlike the previous work, which addresses 3D object detection and semantic segmentation as separate tasks, we extend the model to perform 3D semantic segmentation in addition to 3D object detection.
By combining both tasks into a single model, we are able to better utilize compute resources available on a self-driving vehicle.
Our approach can be trained end-to-end without requiring additional 2D image labels.
On a large dataset, we achieve state-of-the-art performance on both 3D object detection and semantic segmentation tasks.
Figure \ref{fig:example} shows an example of our input data and resulting predictions.

\begin{figure*}[t]
\centering
\includegraphics[width=0.95\textwidth]{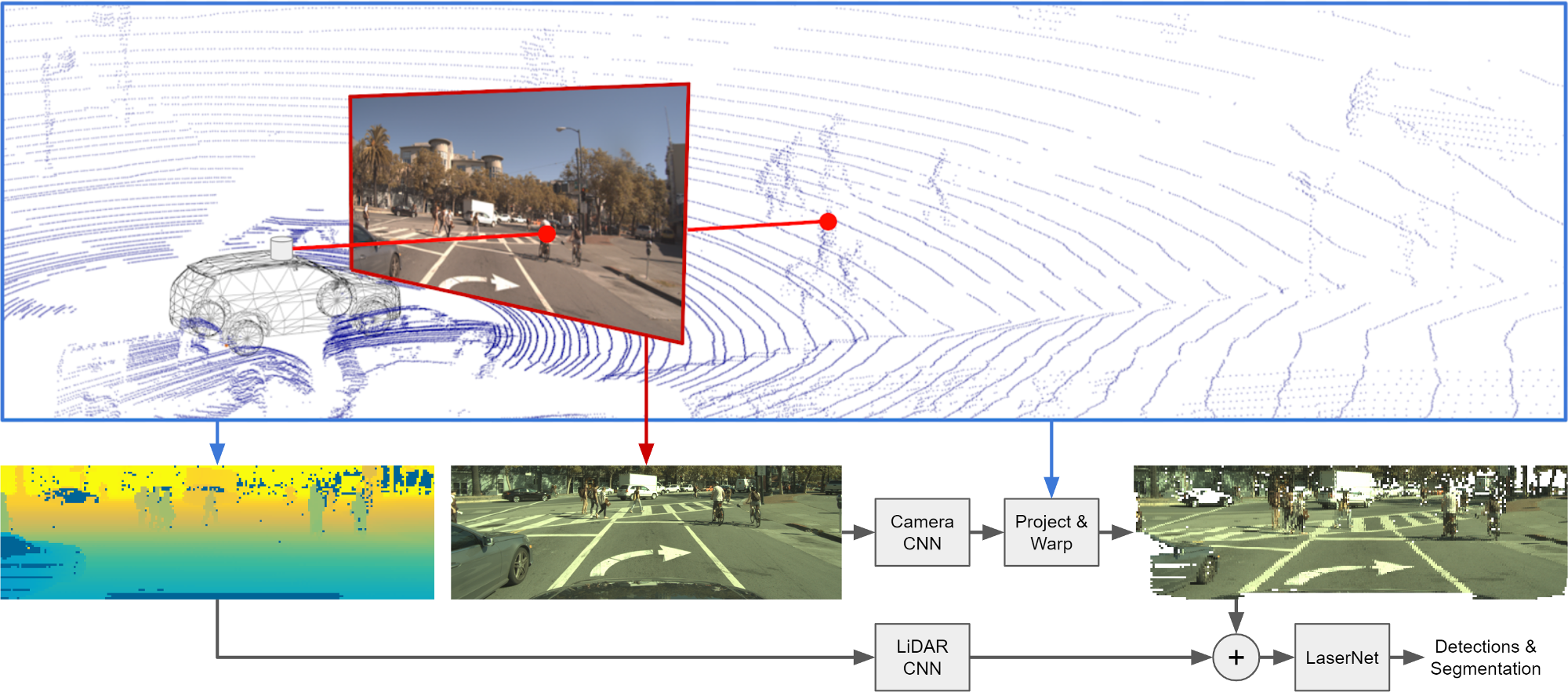}
\caption{Our proposed method fuses 2D camera images and 3D LiDAR measurements to improve 3D object detection and semantic segmentation. Both sensor modalities are represented as images, specifically the 3D data is represented using the native range view of the LiDAR (Section \ref{sec:input}). Our approach associates LiDAR points with camera pixels by projecting the 3D points onto the 2D image, and this mapping is used to warp information from the camera image to the LiDAR image (Section \ref{sec:sensor_fusion}). Instead of warping RGB values as depicted, we fuse features extracted by a CNN (Section \ref{sec:network}). Afterwards, the LiDAR and camera features are concatenated and passed to LaserNet \cite{lasernet}, and the entire model is trained end-to-end to perform 3D object detection and semantic segmentation (Section \ref{sec:predictions}).}
\label{fig:overview}
\end{figure*}

\section{Related Work}
\subsection{3D Object Detection}
Several approaches have been proposed for 3D object detection in the context of autonomous driving. Since LiDAR directly provides range measurements from the surface of objects, it is one of the most popular sensors used for this task \cite{pointpillars,velofcn,lasernet,hdnet,pixor,voxelnet}. Multiple works \cite{mv3d,avod,ibm} have shown that fusing LiDAR data with RGB images improves the performance of object detection especially at long range and for small objects. Thus, in this work, we focus on fusing camera images with a state-of-the-art LiDAR based detector~\cite{lasernet}.

The existing camera and LiDAR fusion methods can be divided into three different groups: \textit{2D to 3D}, \textit{proposal fusion}, and \textit{dense fusion}. In \textit{2D to 3D} approaches~\cite{du2018general,cvpr17mousavian,frustum_pointnet,pointfusion}, 2D object detection is first performed on the RGB images using methods such as \cite{ssd,faster_rcnn}. Afterwards, these 2D boxes are converted to 3D boxes using the LiDAR data. These methods rely on computationally expensive CNNs for 2D detection, and they do not leverage the 3D data to identify objects. Whereas, our approach uses a lightweight CNN to extract features from the 2D image, and we use these features to enrich the 3D data.
Furthermore, these types of methods require both 2D and 3D bounding box labels, while our method only requires 3D labels.

Like the \textit{2D to 3D} approaches, \textit{proposal fusion} methods \cite{mv3d, avod} also consist of two stages. First, they propose 3D bounding boxes by either sampling uniformly over the output space \cite{avod} or by predicting them from the LiDAR data \cite{mv3d}. The 3D proposals are used to extract and combine features from both sensor modalities. The features are extracted by projecting the proposals into each view and pooling over the area encapsulated by the proposal. The features from each sensor are combined and used to produce the final 3D detection. Since these methods require feature pooling for every proposal, they typically have a high runtime.

With \textit{dense fusion} \cite{ibm}, LiDAR and image features are fused into a common frame, which enables single stage 3D object detection. Specifically, \cite{ibm} uses the 3D points from the LiDAR to project features from the image into 3D space, and they use continuous convolutions \cite{wang2018deep} to merge these features into the voxelized BEV. To utilize continuous convolutions, \cite{ibm} needs to identify the $k$ nearest 3D points for each voxel which is a computationally expensive operation. Our proposed method falls into this group; however, we use the 3D points to project the image features directly into the native range view of the LiDAR sensor. We demonstrate that our approach is efficient and can significantly improve detection performance.

\subsection{3D Semantic Segmentation}
Previous work on 3D semantic segmentation has represented LiDAR data in multiple ways: as a point cloud \cite{dohan2015learning,douillard2011segmentation,pointnet++,wang2018deep}, a voxelized 3D space \cite{huang2016point,octnet,tchapmi2017segcloud, zhang2018efficient}, and a spherical image \cite{squeezeseg}.
The accuracy and efficiency of a method depends on its representation of the data.
Methods that discretize the 3D space can lose information through quantization errors, which limits their ability to produce a fine-grain segmentation of the data.
Whereas, methods that operate directly on point cloud data are often slower, since the unstructured nature of the representation does not allow for efficient convolutions.

Fusion of color and geometric data has been extensively explored with the data obtained from RGB-D cameras of indoor scenes \cite{armeni2017joint,silberman2012indoor}. In this setup, the image pixels have both RGB values and a depth measurement. Methods which represent this data as a 3D point cloud \cite{landrieu2018large, tchapmi2017segcloud} decorate each point with its corresponding RGB values and feed it as input to their model. Methods which represent the RGB-D data as separate images  \cite{gupta2014learning,hazirbas2016fusenet,park2017rdfnet,wang2016learning}, extract features from both images and combine the features at multiple scales using a CNN. Since there is a dense correspondence between RGB and depth, the fusion is performed at a per-pixel level.
For outdoors scene with LiDAR, there is only a sparse correspondence between the camera pixels and the range measurements. We demonstrate that combining the RGB values with their corresponding LiDAR point does not help performance. Alternatively, we extract features from the RGB image using a CNN and then fuse those features with the LiDAR points in the native range view of the sensor.
We show that at long ranges, the additional image data improves the semantic segmentation of the 3D points.

\section{Proposed Method}
In the following sections, we describe our modification to LaserNet \cite{lasernet} to fuse RGB image data, and to jointly perform semantic segmentation of the 3D point cloud in addition to 3D object detection.
An overview of our proposed method can be seen in Figure \ref{fig:overview}.

\subsection{Input Data}
\label{sec:input}
Self-driving vehicles leverage a suite of sensors to collect data from its environment.
The input to our proposed method is 3D data from a Velodyne 64E LiDAR, and 2D data from a RGB camera.
An example of the input data is shown in Figure \ref{fig:overview}.

The Velodyne 64E LiDAR measures the surrounding geometry by sweeping over the scene with a set of 64 lasers.
For each measurement, the sensor provides a range $r$, reflectance $e$, azimuth angle $\theta$ of the sensor, and elevation angle $\phi$ of the laser that generated the return.
The 3D position of the measurement can be computed as follows:
\begin{equation}
\boldsymbol{p} =
\begin{bmatrix}
x \\ y \\ z
\end{bmatrix} =
\begin{bmatrix}
r \cos\phi \cos\theta \\
r \cos\phi \sin\theta\\
r \sin\phi
\end{bmatrix}
\end{equation}
where $\boldsymbol{p}$ is ordinarily referred to as a LiDAR point.
As in \cite{lasernet}, we form an image by mapping lasers to rows and discretizing the azimuth angle into columns.
For each cell in the image that contains a measurement, we populate a set of channels with the LiDAR point's range $r$, height $z$, azimuth angle $\theta$, and intensity $e$, as well as a flag indicating if the cell is occupied.
The result is a five channel LiDAR image.

The camera captures a RGB image which covers the front $90^\circ$ horizontal and the full $30^\circ$ vertical field of view of the LiDAR image.
We crop both the RGB and LiDAR image to align the field of views of the sensors, which results in a $512\times64\times5$ LiDAR image and a $1920\times640\times3$ RGB image.
These two images are the input to our model.

\subsection{Sensor Fusion}
\label{sec:sensor_fusion}

As illustrated in Figure \ref{fig:overview}, the 2D image and the 3D points are related through projective geometry.
To fuse the LiDAR and RGB data, we begin by projecting each LiDAR point $\boldsymbol{p}$ onto the RGB image,
\begin{equation}
\alpha \left[ u, v, 1 \right]^T = \boldsymbol{K} \left( \boldsymbol{R} \boldsymbol{p} + \boldsymbol{t} \right)
\end{equation}
where $(u, v)$ is the pixel coordinate of the 3D point in the 2D image, $\boldsymbol{K}$ is the intrinsic calibration matrix of the camera, and $\boldsymbol{R}$ and $\boldsymbol{t}$ are the rotation matrix and translation vector that transform the 3D point from the LiDAR's coordinate frame to the camera's coordinate frame.
As a result, we obtain a mapping from the LiDAR image to the RGB image, and we can use this mapping to copy features from the RGB image into the LiDAR image, as shown in Figure \ref{fig:overview}.
If we fuse raw RGB data in this way, a significant amount of information would be discarded.
Alternatively, we can fuse learned features extracted by a CNN from the RGB image. This allows the network to capture higher level concepts from the image data, so that more information is conveyed when fused with the LiDAR image. The CNN used by our method to extract features from the RGB image is described in Section \ref{sec:network}.

If the feature map has a different resolution than the original image, we update the mapping between points and pixel by dividing the pixel coordinate by the difference in scale $(u/s_x, v/s_y)$.
Afterwards, the pixel coordinates are rounded to the nearest integer value.

Although we only use a single camera in this paper, it is straightforward to extend this approach to incorporate multiple cameras.

\subsection{Network Architecture}
\label{sec:network}
Our network architecture consists of two main components: an auxiliary network used to extract features from the RGB image, and a primary network designed to process features from both sensors.

\begin{figure}
\centering
\includegraphics[width=0.3\textwidth]{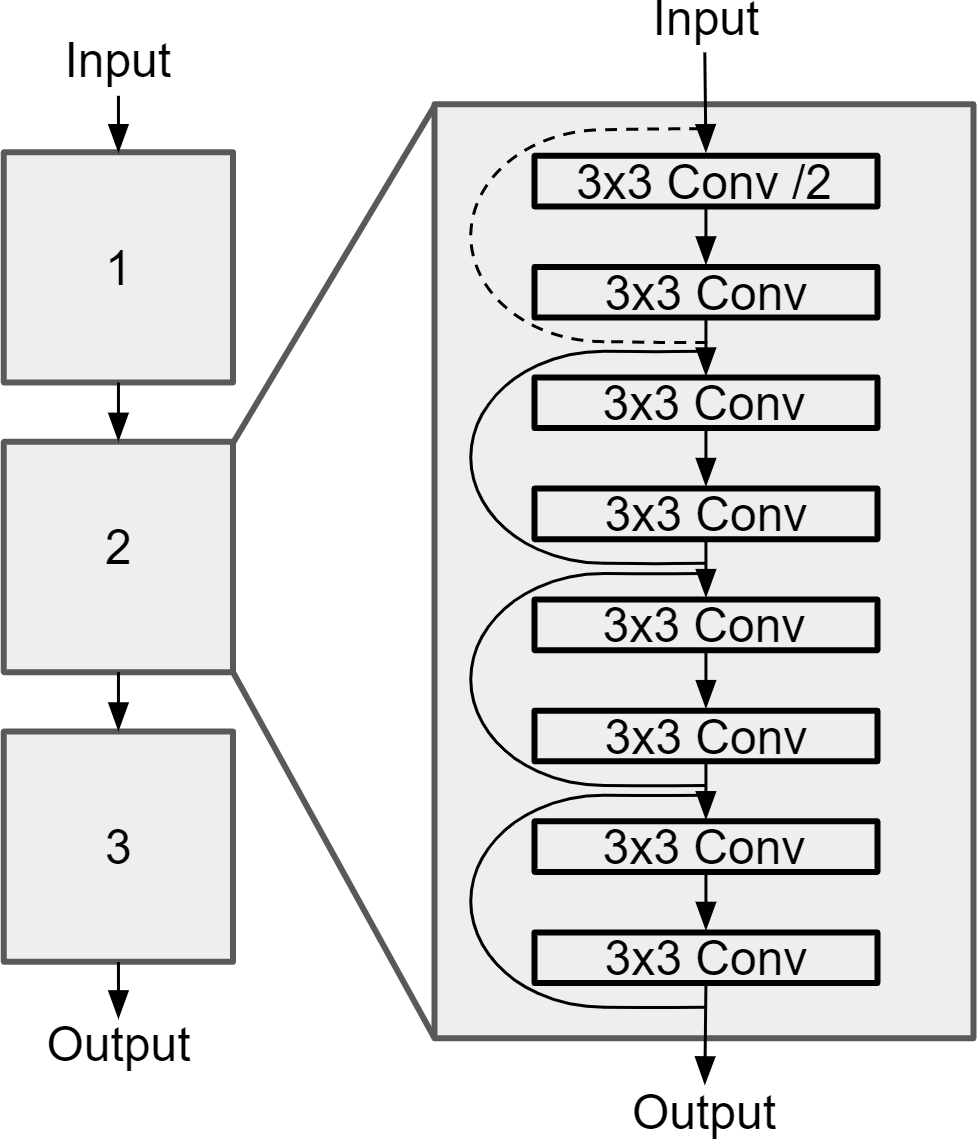}
\caption{Our network used to extract image features (left), which is constructed from a set of ResNet blocks (right). The dashed line implies a convolution is used to reshape the feature map.}
\label{fig:image_network}
\end{figure}

The auxiliary network, shown in Figure \ref{fig:image_network}, takes a RGB image as input and produces a feature map.
This network contains three ResNet blocks \cite{resnet}, where each block downsamples the feature map by half and performs a set of 2D convolutions.
The number of the feature channels in each block is 16, 24, and 32 respectively.

The features extracted from the RGB image are warped into the LiDAR image using the method described in Section \ref{sec:sensor_fusion}.
If there is no valid mapping between a point in the LiDAR image and a pixel the RGB image, the image feature vector at that position is set to all zeros.
Afterwards, the LiDAR image contains a set of feature channels derived directly from the LiDAR data, as well as, a set of feature channels extracted and warped from the RGB image.
To ensure both sensors contribute to the same number of channels, we expand the LiDAR feature channels by passing them through a single $3\times3$ convolutional layer.

The channels from both sensors are concatenated and passed to the primary network.
The deep layer aggregation network \cite{dla} described in \cite{lasernet} is used as our primary network.
Finally, a $1\times1$ convolution is used to convert the output of the network into our predictions.

\subsection{Predictions and Training}
\label{sec:predictions}
In the previous work \cite{lasernet}, the model is trained to predict a set of class probabilities and a set of bounding boxes for each point in the LiDAR image.
Since the model classifies LiDAR points as vehicle, bike, pedestrian, or background, it already performs 3D semantic segmentation to some extent.
To provide more information to downstream components in a self-driving system, we increase the number of class to distinguish between background and road as well as bicycles and motorcycles.

The training procedure is mostly unchanged from \cite{lasernet}.
We simply add the additional classes to the classification loss, and we do not modify the regression loss.
Although the loss is applied at each point in the LiDAR image, the parameters of the auxiliary network can be updated by back-propagating the loss through the projected image features.
It is important to note that the image feature extractor requires no additional supervision; therefore, no supplemental 2D image labels are necessary.

\section{Experiments}
\label{sec:experiments}

\begin{table*}[t]
\centering
\caption{BEV Object Detection Performance}
\vspace{-0.25em}
\scalebox{0.75}
{
    \begin{tabular}{c|c|cccc|cccc|cccc}
    \hline
    \multirow{2}{*}{Method} & \multirow{2}{*}{Input} & \multicolumn{4}{c|}{Vehicle $AP_{0.7}$ }& \multicolumn{4}{c|}{Bike $AP_{0.5}$} & \multicolumn{4}{c}{Pedestrian $AP_{0.5}$}\\
    & & 0-70m & 0-30m & 30-50m & 50-70m & 0-70m & 0-30m & 30-50m & 50-70m & 0-70m & 0-30m & 30-50m & 50-70m\\
    \hline
    PIXOR ~\cite{pixor} & LiDAR  & 80.99  & 93.34  & 80.20 & 60.19  & - & - & -  & - & - & - & - & -\\
    PIXOR++ ~\cite{hdnet} & LiDAR  & 82.63  & 93.80 & 82.34  & 63.42  & - & - & -  & - & - & - & - & -\\
    ContFuse ~\cite{ibm} & LiDAR  & 83.13  & 93.08  & 82.48  & 65.53  & 57.27  & 68.08  & 48.83 & 38.26  & 73.51  & 80.60 & 71.68  & 59.12\\
    LaserNet \cite{lasernet} & LiDAR  & 85.34 & \textbf{95.02} & 84.42 & 67.65  & 61.93 & 74.62 & 51.37 & 40.95  & 80.37 & 88.02 & 77.85 & 65.75\\
    \hline
    ContFuse ~\cite{ibm} & LiDAR+RGB  & 85.17  & 93.86  & 84.41  & 69.83 & 61.13  & 72.01  & 52.60 & 43.03 & 76.84 & 82.97 & 75.54 & 64.19\\
    LaserNet++ (Ours) & LiDAR+RGB  & \textbf{86.23} & 94.96 & \textbf{85.42} & \textbf{70.31}  & \textbf{65.68} & \textbf{76.36} & \textbf{56.52} & \textbf{50.08} & \textbf{83.42} & \textbf{91.12} & \textbf{81.43} & \textbf{70.97}\\
    \hline
    \end{tabular}
}
\label{tab:car3d_detection_results}
\end{table*}

\begin{table*}[t]
\centering
\caption{3D Semantic Segmentation Performance}
\vspace{-0.25em}
\scalebox{0.75}
{
    \begin{tabular}{c|c|cc|cccccc}
    \hline
    \multirow{2}{*}{Method} & \multirow{2}{*}{Input} & \multirow{2}{*}{mAcc} & \multirow{2}{*}{mIoU} & \multicolumn{6}{c}{Class IoU}\\
    & & & & Background & Road & Vehicle & Pedestrian & Bicycle & Motorcycle \\
    \hline
    2D U-Net \cite{zhang2018efficient} & LiDAR & 81.95 & 76.39 & 92.03 & 97.92 & 93.76 & 74.47 & 61.25 & 38.90 \\
    LaserNet++ (Ours) & LiDAR+RGB & \textbf{91.77} & \textbf{86.62} & \textbf{93.59} & \textbf{98.23} & \textbf{97.67} & \textbf{86.19} & \textbf{80.98} & \textbf{63.07} \\
    \hline
    \end{tabular}
}
\label{tab:car3d_segmentation_results}
\end{table*}

\begin{table}[t]
\centering
\caption{Ablation Study for Semantic Segmentation}
\vspace{-0.25em}
\scalebox{0.75}
{
    \begin{tabular}{c|cccc}
    \hline
    \multirow{2}{*}{Image Features} & \multicolumn{4}{c}{mIoU} \\
    & 0-70m & 0-30m & 30-50m & 50-70m \\
    \hline
    None & 86.37 & 87.51 & 74.89 & 64.38 \\
    RGB Features & 86.60 & \textbf{87.70} & 75.09 & 64.75 \\
    CNN Features & \textbf{86.62} & 87.59 & \textbf{76.05} & \textbf{69.57} \\
    \hline
    \end{tabular}
}
\label{tab:car3d_segmentation_ablation_study}
\end{table}

Our proposed method is evaluated and compared to state-of-the-art methods in both 3D object detection and semantic segmentation on the large-scale ATG4D dataset.
The dataset contains a training set with 5,000 sequences sampled at 10 Hz for a total of 1.2 million images.
The validation set contains 500 sequences sampled at 0.5 Hz for a total of 5,969 images.
We evaluate the detections and segmentation within the front $90^\circ$ field of view and up to $70$ meters away.

To train the network, we use the settings described in \cite{lasernet}.
We train for 300k iterations with a batch size of 128 distributed over 32 GPUs.
The learning rate is initialized to 0.002 and decayed exponentially by 0.99 every 150 iterations.
Furthermore, we utilize the Adam optimizer \cite{adam}.

\subsection{3D Object Detection}
The performance of our approach and existing state-of-the-art methods on the task of 3D object detection is shown in Table \ref{tab:car3d_detection_results}.
Following the previous work, we use the average precision (AP) metric.
To be considered a true positive, a detection must achieve a significant intersection-over-union (IoU) with the ground truth.
For vehicles, we require a 0.7 IoU, and for pedestrians and bikes, we use a 0.5 IoU.
The existing detectors do not differentiate between bicycles and motorcycles, so for comparisons, we merge the two classes into a single bike class.

In most cases, our proposed method out-performs the existing state-of-the-art methods on this dataset.
Compared to methods that solely utilize LiDAR data, our approach does significantly better at longer ranges.
The LiDAR measurements are reasonably dense in the near range, but fairly sparse at long range.
Adding the supplemental 2D data improves performance where the 3D data is scarce; conversely, less benefit is observed where the 3D data is abundant.

On smaller objects (pedestrian and bike), our approach significantly out-performs the existing method that uses both LiDAR and RGB data.
We believe this is due to our method representing the LiDAR data using a RV where the previous work uses a BEV representation \cite{ibm}.
Unlike the RV, the BEV requires the 3D data to be voxelized, which results in fine-grain detail being removed.

\subsection{3D Semantic Segmentation}
The evaluation of our proposed method on the task of 3D semantic segmentation compared to the existing state-of-the-art is shown in Table \ref{tab:car3d_segmentation_results}.
To assess the methods, we use the mean class accuracy (mAcc), the mean class IoU (mIoU), and the per-class IoU computed over the LiDAR points as defined in \cite{zhang2018efficient}.
To perform semantic segmentation, we classify each point in the LiDAR image with its most likely class according to the predicted class probabilities.
If more than one point fall into the same cell in the LiDAR image, only the closest point is classified, and the remaining points are set to an unknown class.
Since the resolution of the image is approximately the resolution of the LiDAR, it is uncommon for multiple points to occupy the same cell.
For comparisons, we implement the method proposed in \cite{zhang2018efficient}, and we incorporate focal loss \cite{retinanet} into their method to improve performance.

On this dataset, our approach considerably out-performs this state-of-the-art method across all metrics.
It performs particularly well on smaller classes (pedestrian, bicycle, and motorcycle).
Again, we believe this is due to our approach using a RV instead of the BEV representation used in the previous work \cite{zhang2018efficient}.
The BEV voxelizes the 3D points, so precise segmentation of small objects is challenging.

\begin{figure}
\centering
\includegraphics[width=0.45\textwidth]{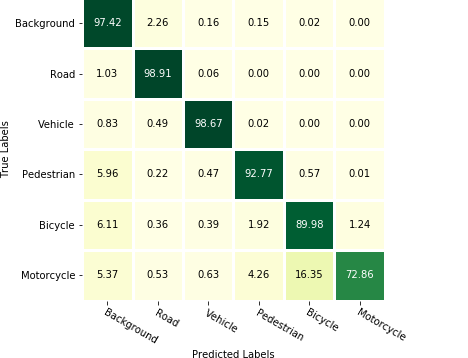}
\caption{The confusion matrix for our method on the task of 3D semantic segmentation.}
\label{fig:confusion_matrix}
\end{figure}

In Table \ref{tab:car3d_segmentation_ablation_study}, we study the effect of different image features on semantic segmentation.
Since the LiDAR data becomes sparse at far ranges, the segmentation metrics are dominated by the near range performance.
We know from Table \ref{tab:car3d_detection_results} that image features improve long range performance; therefore, we examine the segmentation performance at multiple ranges.
In the near range, there is practically no benefit from fusing image features.
However, at long range, fusing image features extracted by a CNN considerably improves performance.
Fusing raw RGB values has little effect on performance.
Lastly, Figure \ref{fig:confusion_matrix} shows the confusion matrix for our approach.
Unsurprisingly, the majority of confusion is between the motorcycle and bicycle class.

\begin{figure*}
\vspace{3.5em}
\begin{subfigure}{\textwidth}
  \centering
  \includegraphics[width=0.9\linewidth]{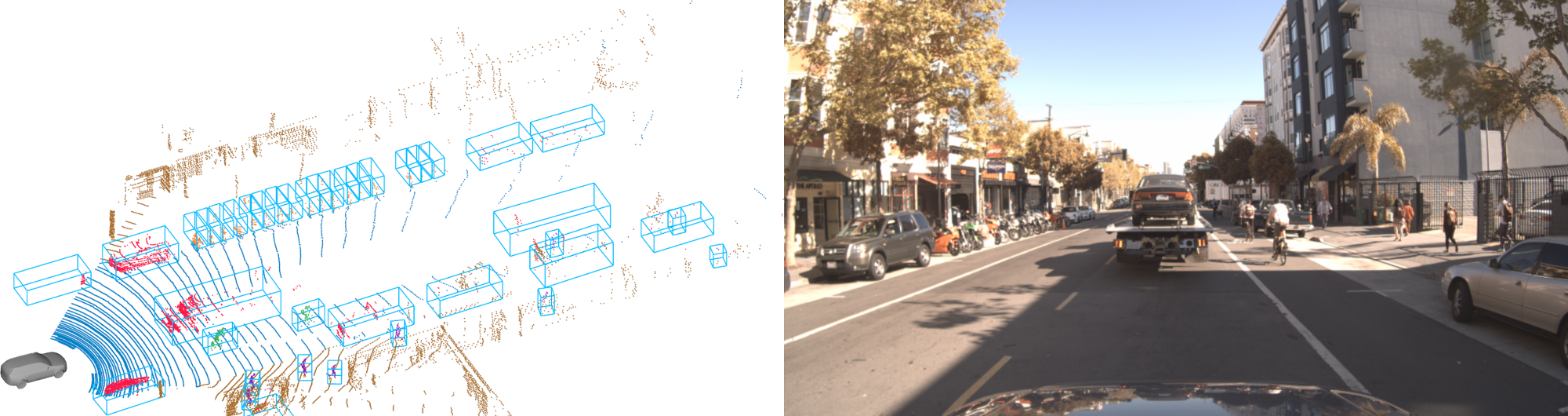}
  \label{fig:qual1}
  \vspace{0.5em}
\end{subfigure}
\begin{subfigure}{\textwidth}
  \centering
  \includegraphics[width=.9\linewidth]{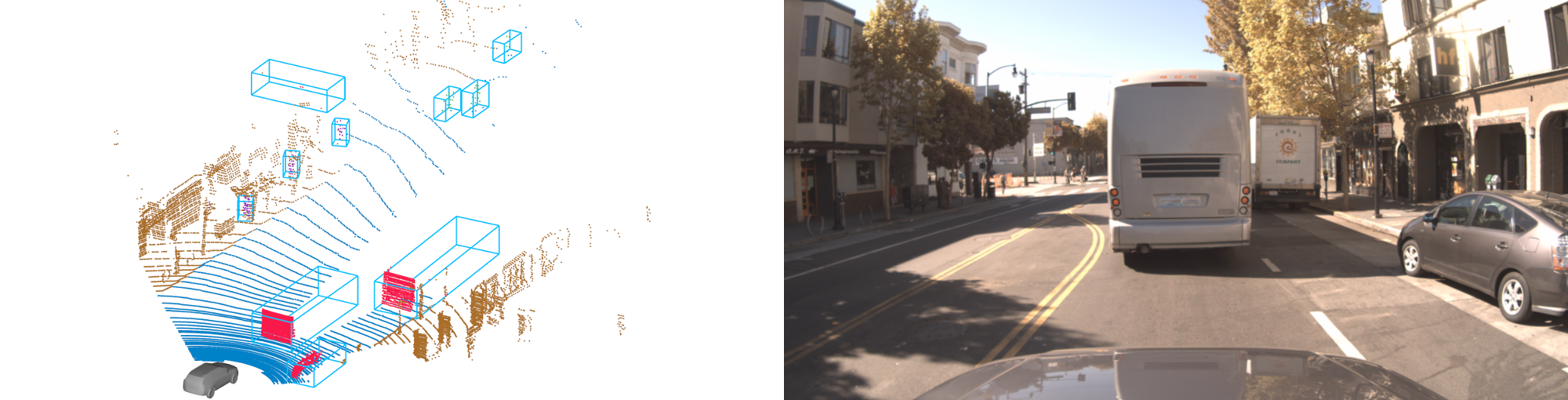}
  \label{fig:qual2}
  \vspace{0.5em}
\end{subfigure}
\begin{subfigure}{\textwidth}
  \centering
  \includegraphics[width=.9\linewidth]{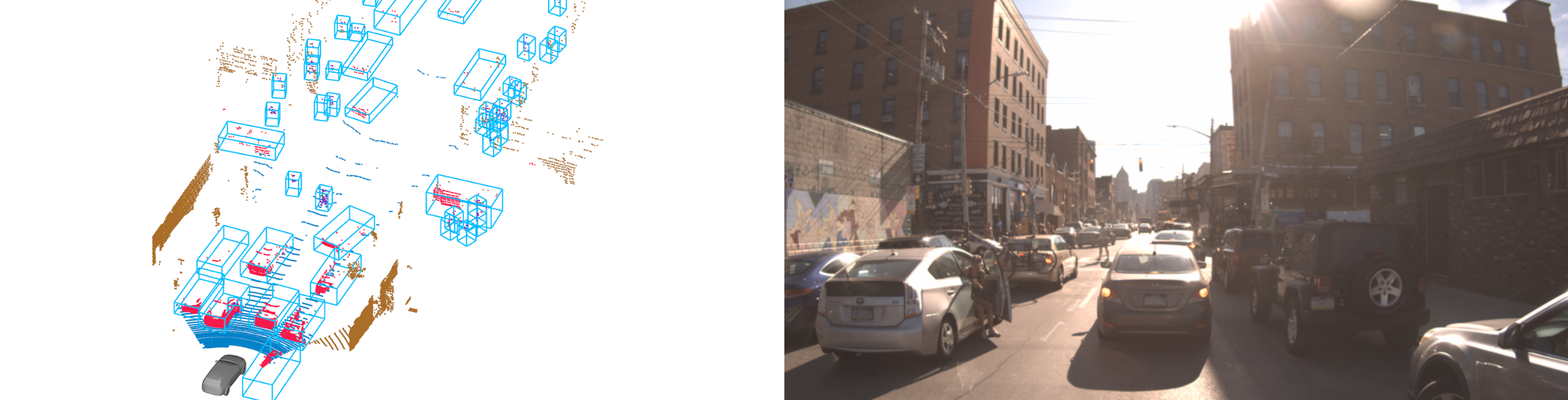}
  \label{fig:qual4}
  \vspace{0.5em}
\end{subfigure}
\begin{subfigure}{\textwidth}
  \centering
  \includegraphics[width=.9\linewidth]{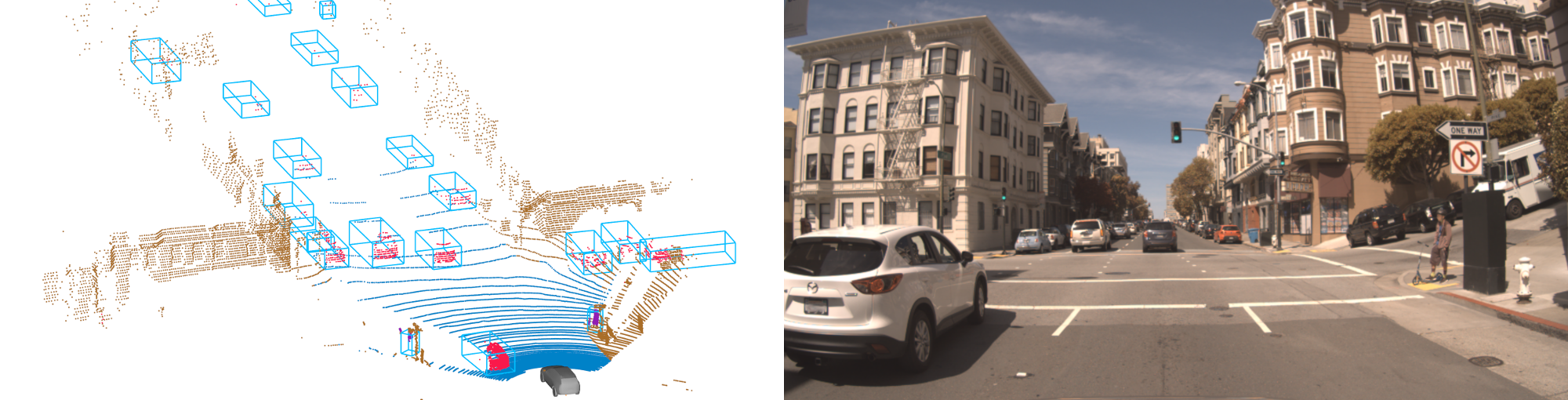}
  \label{fig:qual3}
  \vspace{0.5em}
\end{subfigure}
\begin{subfigure}{\textwidth}
  \centering
  \includegraphics[width=.4\linewidth]{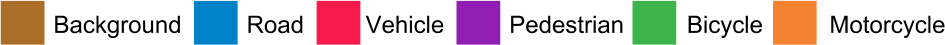}
  \label{fig:qual_legend}
\end{subfigure}
\caption{A few interesting successes and failures of our proposed method. (Top) Our approach is able to detect every motorcycle in a large row of parked motorcycles. (Second) Our method is able to detect several bikes which are approximately 50 to 60 meters away from the self-driving vehicle where LiDAR is very sparse. (Third) The network classifies most of the LiDAR points on the person getting out of a car as vehicle, however it still produces the correct bounding box. This is a benefit of predicting bounding boxes at every LiDAR point. (Bottom) Due to the steep elevation change in the road on the right side, the model incorrectly predicts the road points as background.}
\label{fig:qualitative}
\vspace{3.5em}
\end{figure*}

Figure \ref{fig:qualitative} shows qualitative results for our method on both tasks, 3D object detection and 3D semantic segmentation.

\subsection{Runtime Evaluation}
Runtime performance is critical in a full self-driving system.
LaserNet \cite{lasernet} was proposed as an efficient 3D object detector, and our extensions are designed to be lightweight.
As shown in Table \ref{tab:run_time_results}, the image fusion and the addition of semantic segmentation only adds 8 ms (measured on a NVIDIA TITAN Xp GPU). Therefore, our method can detect objects and perform semantic segmentation at a rate greater than 25 Hz.

\begin{table}[t]
\centering
\caption{Runtime Performance}
\vspace{-0.25em}
\scalebox{0.75}{
  \begin{tabular}{ c|cc }
    \hline
    Method &  Forward Pass (ms) & Total (ms) \\
    \hline
    LaserNet \cite{lasernet} & 12 & 30 \\
    LaserNet++ (Ours) & 18 & 38 \\
    \hline
  \end{tabular}
}
\label{tab:run_time_results}
\end{table}

\section{Conclusion}
In this work, we present an extension to LaserNet \cite{lasernet} to fuse 2D camera data with the existing 3D LiDAR data, achieving state-of-the-art performance in both 3D object detection and semantic segmentation on a large dataset. Our approach to sensor fusion is straightforward and efficient. Also, our method can be trained end-to-end without any 2D labels. The addition of RGB image data improves the performance of the model, especially at long ranges where LiDAR measurements are sparse and on smaller objects such as pedestrians and bikes.

Additionally, we expand the number of semantic classes identified by the model, which provides more information to downstream components in a full self-driving system. By combining both tasks into a single network, we reduce the compute and latency that would occur by running multiple independent models.

\section{Acknowledgements}
Both LaserNet and LaserNet++ would not be possible without the help of countless members of the Uber \mbox{Advanced Technologies Group}. In particular, we would like to acknowledge the labeling team, who build and maintain large-scale datasets like the ATG4D dataset.

{\small
\bibliographystyle{ieee_fullname}
\bibliography{egbib}
}
\end{document}